\documentclass{article}
\usepackage{ijcai23}
\usepackage{makecell}
\usepackage{times}
\usepackage{soul}
\usepackage{url}
\usepackage[hidelinks]{hyperref}
\usepackage[utf8]{inputenc}
\usepackage[small]{caption}
\usepackage{graphicx}
\usepackage{amsmath}
\usepackage{amsthm}
\usepackage{booktabs}
\usepackage[switch]{lineno}
\usepackage[ruled]{algorithm2e}
\usepackage{algpseudocode}
\usepackage[bottom]{footmisc}
\usepackage{floatrow}
\newfloatcommand{capbtabbox}{table}[][\FBwidth]

\title{Enhancing Efficient Continual Learning with Dynamic Structure Development of Spiking Neural Networks}

\author{
Bing Han$^{1,2}$\and
Feifei Zhao$^1$\and
Yi Zeng$^{1,2,3,4}$\footnotemark[1]\and
Wenxuan Pan $^{1,2}$\And
Guobin Shen $^{1,3}$\\
\affiliations
$^1$Brain-inspired Cognitive Intelligence Lab, Institute of Automation, Chinese Academy of Sciences\\
$^2$ School of Artificial Intelligence, University of Chinese Academy of Sciences\\
$^3$ School of Future Technology, University of Chinese Academy of Sciences\\
$^4$Center for Excellence in Brain Science and Intelligence Technology, Chinese Academy of Sciences
\emails 
\{hanbing2021, zhaofeifei2014, yi.zeng, panwenxuan2020, shenguobin2021\}@ia.ac.cn
}

\begin{document}
\maketitle
\renewcommand{\thefootnote}{\fnsymbol{footnote}}
\footnotetext[1]{Corresponding author.}

\begin{abstract}
Children possess the ability to learn multiple cognitive tasks sequentially, which is a major challenge toward the long-term goal of artificial general intelligence. Existing continual learning frameworks are usually applicable to Deep Neural Networks (DNNs) and lack the exploration on more brain-inspired, energy-efficient Spiking Neural Networks (SNNs). Drawing on continual learning mechanisms during child growth and development, we propose Dynamic Structure Development of Spiking Neural Networks (DSD-SNN) for efficient and adaptive continual learning. When learning a sequence of tasks, the DSD-SNN dynamically assigns and grows new neurons to new tasks and prunes redundant neurons, thereby increasing memory capacity and reducing computational overhead.
In addition, the overlapping shared structure helps to quickly leverage all acquired knowledge to new tasks, empowering a single network capable of supporting multiple incremental tasks (without the separate sub-network mask for each task). We validate the effectiveness of the proposed model on multiple class incremental learning and task incremental learning benchmarks. Extensive experiments demonstrated that our model could significantly improve performance, learning speed and memory capacity, and reduce computational overhead. Besides, our DSD-SNN model achieves comparable performance with the DNNs-based methods, and significantly outperforms the state-of-the-art (SOTA) performance for existing SNNs-based continual learning methods.
\end{abstract}

\section{Introduction}
Children are able to incrementally learn new tasks to acquire new knowledge, however, this is a major challenge for Deep Neural Networks (DNNs) and Spiking Neural Networks (SNNs). When learning a series of different tasks sequentially, DNNs and SNNs forget the previously acquired knowledge and fall into catastrophic forgetting~\cite{french1999catastrophic}. Despite some preliminary solutions that have recently been proposed for DNNs-based continual learning, there is still a lack of in-depth inspiration from brain continual learning mechanisms and exploration on SNNs-based models. 

The studies attempt to address the continual learning problem of DNNs under task incremental learning (recognition within the classes of a known task) and class incremental learning (recognition within all learned classes) scenarios. Related works can be roughly divided into three categories:

 \paragraph{a) Regularization.} Employing maximum a posterior estimation minimizes the changes of important weights~\cite{li2017learning,kirkpatrick2017overcoming,zenke2017continual}. These methods require strong model assumptions, such as the EWC~\cite{kirkpatrick2017overcoming} supposing that new weights are updated to local regions of the previous task weights, which are highly mathematical abstractions and poorly biologically plausibility. 
 
  \paragraph{b) Replay and retrospection.} Reviewing a portion of the samples of the old tasks while learning the new task~\cite{lopez2017gradient,van2020brain,kemker2017fearnet}, is currently considered as the superior class incremental learning method. The samples of old tasks are stored in additional memory space or generated by additional generation networks, resulting in extra consumption. 
 
 \paragraph{c) Dynamic network structure expansion. }\cite{rusu2016progressive,siddiqui2021progressive} proposed progressive neural networks that extend a new network for each task, causing a linear increase in network scale. To reduce network consumption, a sub-network of the whole is selected for each task using pruning and growth algorithms~\cite{yoon2017lifelong,dekhovich2022continual}, evolutionary algorithms~\cite{fernando2017pathnet} or reinforcement learning (RL) algorithms~\cite{xu2018reinforced,gao2022efficient}. However, these methods require storing a mask for each sub-network, which to some extent amounts to storing a separate network for each task, rather than a brain-inspired overall network capable of performing multiple sequential tasks simultaneously. 

To the best of our knowledge, there is little research on SNNs-based continual learning. Spiking neural networks, as third-generation neural networks~\cite{Maass1997Networks,zhao2022nature}, simulate the information processing mechanisms of the brain, and thus serve well as an appropriate level of abstraction for integrating inspirations from brain multi-scale biological plasticity to achieve child-like continual learning. The existing HMN algorithm~\cite{zhao2022framework} uses a DNN network to decide the sub-network of SNN for each task, and is only applicable to two-layer fully connected networks for the N-MNIST dataset. There is still a lack of SNNs-based continual learning methods that could incorporate in-depth inspiration from the brain's continual learning mechanisms, while achieving comparable performance with DNNs under complex continual learning scenarios.

Structural development mechanisms allow the brain's nervous system to dynamically expand and contract, as well as flexibly allocate and invoke neural circuits for efficient continual learning~\cite{silva2009molecular}. Motivated by this, this paper proposes Dynamic Structure Development of Spiking Neural Networks (DSD-SNN) for efficient and adaptive continual learning. DSD-SNN is designed as an SNN architecture that can be dynamically expanded and compressed, empowering a single network to learn multiple incremental tasks simultaneously, overcoming the problem of needing to assign masks to each task faced by DNNs-based continual learning methods.
 We validate the effectiveness of our proposed model on multiple class incremental learning (CIL) and task incremental learning (TIL) benchmarks, achieving comparable or better performance on MNIST, N-MNIST, and CIFAR-100 datasets. Especially, the proposed DSD-SNN model achieves an accuracy of 77.92\% $\pm$ 0.29\% on CIFAR100, only using 37.48\% of the network parameters.

The main contributions of this paper can be summarized as follows:

\begin{itemize}
    \item [$\bullet$]  DSD-SNN dynamically grows new neurons to learn newly arrived tasks, while extremely compressing the network to increase memory capacity and reduce computational overhead.
 
    \item [$\bullet$]
DSD-SNN maximally utilizes the previously learned tasks to help quickly adapt and infer new tasks, enabling efficient and adaptive continual learning (no need to identify separate sub-network mask for each task).
 
    \item [$\bullet$] The experimental results demonstrate the remarkable superiority of DSD-SNN model on performance, learning speed, memory capacity and computational overhead compared with the state-of-the-art (SOTA) SNNs-based continual learning algorithms, and comparable performance with DNNs-based continual learning algorithms.
   
\end{itemize}

\section{Related Work}
This paper mainly focuses on dynamic network structure expansion algorithms based on structural plasticity, which can be divided into progressive neural networks (PNN) and sub-network selection algorithms. In fact, the existing network structure expansion algorithms are mostly DNNs-based continual learning, with little exploration on SNNs.

 \paragraph{Progressive neural networks.} ~\cite{rusu2016progressive} first proposes the progressive neural network and applies it to multiple continual reinforcement learning tasks. The PNN expands a new complete network for each new task and fixes the networks of the old tasks. In addition, lateral connections are introduced between the networks to effectively leverage the knowledge already learned. PIL~\cite{siddiqui2021progressive} extends the PNN to large-scale convolutional neural networks for image classification tasks. However, the PNNs algorithms extremely increase the network storage and computational consumption during continual learning. In contrast, as development matures and cognition improves, the number of brain synapses decreases by more than 50\% ~\cite{huttenlocher1990morphometric}, forming a highly sparse brain structure perfect for continual learning. The PNNs blindly expand the structure causing catastrophic effects in the case of massive sequential tasks.

 \paragraph{Sub-network selection algorithm.} A part of the network nodes is selected to be activated for a given task. PathNet~\cite{fernando2017pathnet} is first proposed to select path nodes (each node contains a set of neurons) for each task using the genetic algorithm. RPS-Net~\cite{rajasegaran2019random} randomly activates multiple input-to-output paths connected by convolutional blocks, and chooses the highest-performing ones as the final path. In addition, RCL~\cite{xu2018reinforced} employ additional RL networks to learn the number of neurons required for a new task, while CLEAS~\cite{gao2022efficient} uses RL to directly determine the activation and death of each neuron. HMN~\cite{zhao2022framework} uses a hybrid network learning framework that uses an ANN modulation network to determine the activation of neurons for a SNN prediction network, but is only applied to small-scale networks for simple scenarios. A sub-network mask learning process based on pruning strategy is proposed by ~\cite{dekhovich2022continual}, which is applied to CIL combined with the replay strategy. The above algorithms select sub-networks for each task separately, failing to maximize the reuse of acquired knowledge to support new task learning.

To solve this problem, DRE~\cite{yan2021dynamically} prunes a sparse convolutional feature extractor for each task, and then merges the output of the convolution extractor into the previous tasks. CLNP~\cite{golkar2019continual} grows new neurons for a new task based on the old network, and DEN~\cite{yoon2017lifelong} expands when the already learned network is insufficient for the new task, while reusing the existing neurons. These several works require storing
an additional sub-network mask for each task, which both increases additional storage consumption and is not consistent with the overall developmental learning process of the brain.

Considering the various limitations of existing works above, the DSD-SNN proposed in this paper, which is a pioneering algorithm on SNNs-based continual learning, enables the capacity of a single network to learn multiple sequential tasks simultaneously, while reusing the acquired knowledge and significantly increasing the memory capacity.

\begin{figure*}[t] 
	\centering  
	\includegraphics[width=0.9\linewidth]{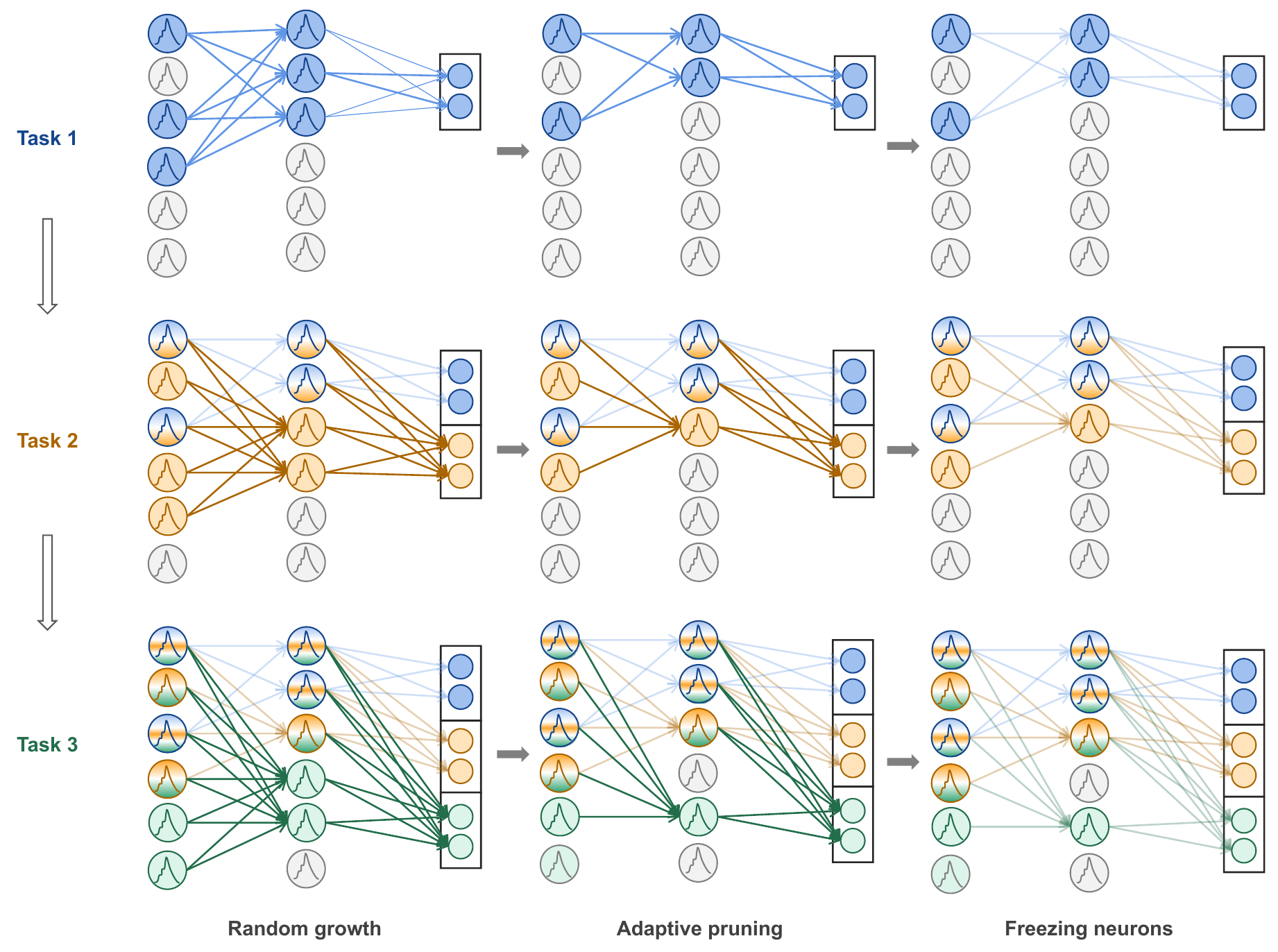} 
	\caption{The DSD-SNN model realizes multi-task incremental learning through random growth, adaptive pruning, and freezing of neurons.}
	\label{tu1}
\end{figure*}

\section{Method}
\subsection{Continual Learning Definition}
We are expected to sequentially learn $\Gamma$ tasks, $\Gamma=\{T_{1},...,T_{N}\}$. Each task $T_{i}$ takes the form of a classification 
problem with its own dataset: $D_{T_{i}}=\{(x_{j},y_{j})\}_{j=1}^{N_{T_i}}$, where $x_j\in \chi, y_i \in\{1,...,C_{T_i}\}$, $\chi$ is the input image space, $N_{T_i}$ and $C_{T_i}$ are the number of samples and classes of task $T_i$.  For the task incremental learning scenario, $T_i$ is knowable in the testing process, setting requires to optimize:
\begin{equation}
    \label{t}
    \mathop{max}\limits_{\theta} E_{T_i\sim \Gamma} [E_{(x_j,y_j)\sim T_i}[log p_{\theta}(y_j|x_j,T_i)]]
\end{equation}
where $\theta$ is the network parameters. When $T_i$ is unknown in testing, more complex class incremental learning scenarios solve the following problems:
\begin{equation}
    \label{c}
    \mathop{max}\limits_{\theta} E_{T_i\sim \Gamma} [E_{(x_j,y_j)\sim T_i}[logp_{\theta}(y_j|x_j)]]
\end{equation}

\subsection{DSD-SNN Architecture}
The design of the DSD-SNN algorithm is inspired by the dynamic allocation, reorganization, growth, and pruning of neurons during efficient continual learning in the brain. As depicted in Fig. \ref{tu1}, the proposed DSD-SNN model includes three modules (random growth, adaptive pruning, freezing neurons) to accomplish multi-task incremental learning. 

 \paragraph{Random growth.} When a new task is coming, the DSD-SNN model first randomly assigns and grows a portion of untrained empty neurons to form a new pathway. And the new task-related classification neurons are added to the output layer as shown in Fig. \ref{tu1}. Newly grown neurons receive the output of all non-empty neurons of the previous layer (both newly grown neurons and already frozen neurons in the previous tasks). Therefore, all features learned from previous tasks can be captured and reused by the neural pathways of the new task. Then, the DSD-SNN algorithm can take full advantage of the features learned from the previous task to help the new task converge quickly, while the newly grown neurons can also focus on learning features specific to the new task.

\begin{figure*}[t] 
	\centering  
	\includegraphics[width=1\linewidth]{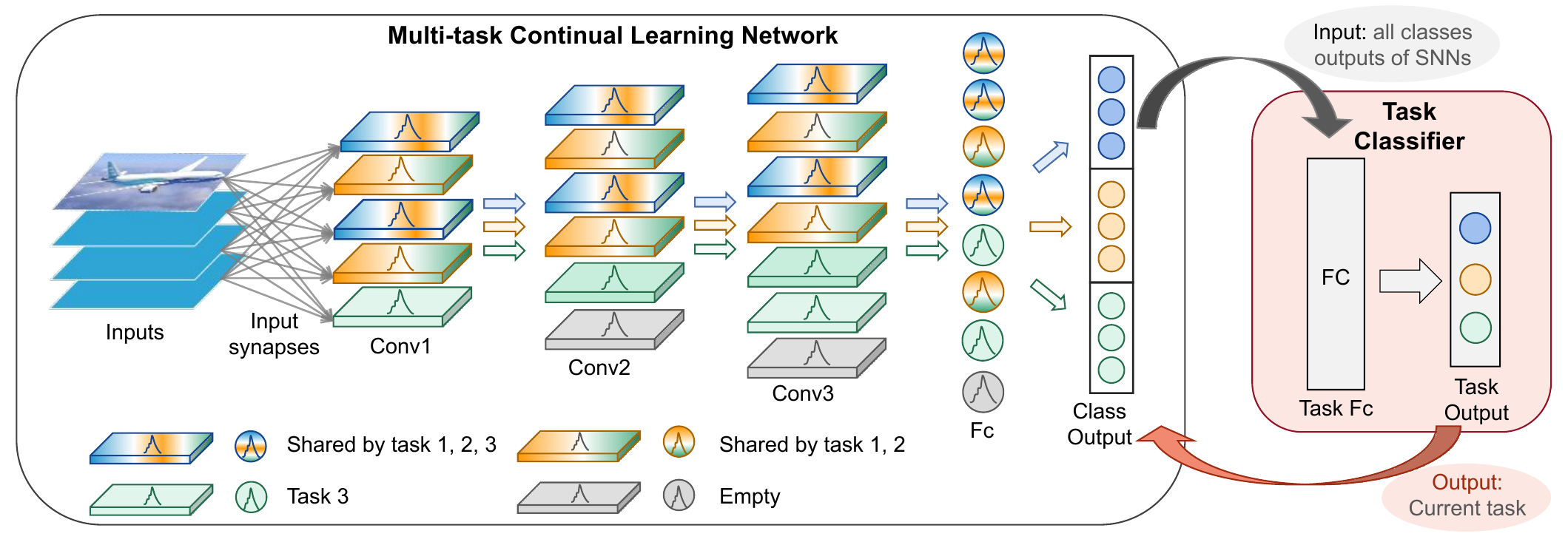} 
	\caption{The architecture of the DSD-SNN model.}
	\label{tu2}
\end{figure*}

 \paragraph{Adaptive pruning.} During the learning process of the current task, the DSD-SNN algorithm adaptively detects relatively inactive neurons in the current pathway based on synaptic activity and prunes those redundant neurons to save resources. The pruned neurons are re-initialized as empty neurons that can be assigned to play a more important role in future tasks. Pruning only targets those neurons that are newly grown for the current task and does not include neurons that were frozen in the previous tasks. Adaptive pruning can substantially expand the memory capacity of the network to learn and memorize more tasks under a fixed scale.

 \paragraph{Freezing neurons.} The contributing neurons that are retained after pruning will be frozen, enabling the DSD-SNN model to learn new tasks without forgetting the old tasks. The frozen neurons can be connected to newly grown neurons to provide acquired knowledge. During the training of new task $T_i$, all input synapses of the frozen neuron are no longer updated, only the newly added output synapses to the new neurons can be updated. The DSD-SNN model with neuron growth, pruning, and freezing can memorize previous knowledge and reuse the acquired knowledge to learn new tasks for efficient continual learning. 

The deep SNN with multiple convolutional and fully connected layers is constructed to implement task incremental learning and class incremental learning, as shown in Fig. \ref{tu2}. During the training process, we sequentially input training samples of each task and update the synapses newly added to the network. In the testing process, test samples of all learned tasks are fed into our overall multi-task continual learning network, so that a single DSD-SNN model can achieve all tasks without the need to identify separate sub-network mask for each task. 

To address more complex class incremental learning, we add a two-layer network as the task classifier. The task classifier receives inputs from the classes outputs of the continual learning network, and outputs which task the current sample belongs to (as in the red box in Fig. \ref{tu2}). According to the inferred task $\hat{T}_i$ obtained from the task classifier, the DSD-SNN model chooses the maximum output class of the $\hat{T}_i$ task in the continual learning network as the predicted class. 

\subsection{DSD-SNN Computational Details}
So far in this section, we have described how our model efficiently and adaptively accomplishes continual learning. We now introduce the detailed growth and pruning scheme that we use throughout this paper.

\subsubsection{Neuronal Growth and Allocation}
During brain development, neurons and synapses are first randomly and excessively grown and then reshaped based on the external experience~\cite{jun2007development,elman1996rethinking}. In the DSD-SNN model, the SNN is first initialized to consist of $N^l$ neurons in each layer $l$. In the beginning, all neurons in the network are unassigned empty neurons $N_{empty}$. When the new task $T_i$ arrives, we randomly grow $\rho\% \times  N^l$ neurons from the empty neurons for each layer, denoted as $N_{new}$. After training and pruning for task $T_i$ , all retained neurons $N_{new}$ are frozen, added to $N_{frozen}$.

To better utilize the acquired knowledge, the newly grown neurons $N_{new}^{l}$ in each layer not only receive the output of the new growth neurons $N_{new}^{l-1}$ in the previous layer, but also receive the output of the frozen neurons $N_{frozen}^{l-1}$ in the previous layer, as follows.
\begin{equation}
    \label{g}
    \{N_{frozen}^{l-1},N_{new}^{l-1}\} \rightarrow N_{new}^{l}
\end{equation}
Where $ \rightarrow $ represents the input connections. For the frozen neurons $N_{frozen}^{l-1}$, growth does not add input connections to avoid interference with the memory of previous tasks.

Note that we do not assign task labels to frozen and new growth neurons in either the training or testing phase of continual learning. That is, the DSD-SNN algorithm uses the entire network containing all neurons that have learned previous tasks to do prediction and inference. Thus, our model is able to learn multiple sequential tasks simultaneously without storing separate sub-network masks.

\subsubsection{Neuronal Pruning and Deactivation}
Neuroscience researches have demonstrated that after the overgrowth in infancy, the brain network undergoes a long pruning process in adolescence, gradually emerging into a delicate and sparse network~\cite{huttenlocher1979synaptic,huttenlocher1990morphometric,zhao2022toward}. Among them, input synapses are important factors to determine the survival of neurons according to the principle of “use it or lose it”~\cite{furber1987naturally,bruer1999neural,zhao2021dynamically}. For SNN, neurons with input synapse weights close to 0 are more difficult to accumulate membrane potentials beyond the spiking threshold, resulting in firing spikes less and contributing to the outputs less. Therefore, we used the sum of input synapse weights $S_i^l$ to assess the importance of neurons $i$ in the $l$ layer as in Eq. \ref{s}.
\begin{equation}
    \label{s}
    S_i^l=\sum_{j=1}^{M_{l-1}} W_{ij}
\end{equation}

Where $W_{ij}$ is the synapse weights from presynaptic neuron $j$ to postsynaptic neuron $i$, $M_{l-1}$ is the number of presynaptic neurons.

During the training of new tasks, we monitor the importance of newly grown neurons $N_{new}$ and prune redundant neurons whose values of $S_i$ are continuously getting smaller. Here, we define a pruning function as follows:
\begin{equation}
	\label{ddd}
	\phi_{P_i^l} =\alpha *Norm(S_i^l)-\rho_p 
\end{equation}

\begin{equation}
	\label{fd}
	P_i^l=\gamma P_i^l +e^{-\frac{epoch}{\eta}} \phi_{P_i^l}
\end{equation}

Where $Norm(S_i^l)$ refers to linearly normalize $S_i^l$ to 0 $\sim $ 1. $\alpha=2$ and $\rho_p$ control the pruning strength. $\rho_p$ includes $\rho_c$ and $\rho_f$ for the convolutional and fully connected layers, respectively. $P_i^l$ is initialized to 5. $\gamma = 0.99$ and $\eta$ controls the update rate as~\cite{han2022developmental}. $e^{-\frac{epoch}{\eta}}$ decreases exponentially with increasing epoch, which is consistent with the speed of the pruning process in biological neural networks that are first fast, then slow, and finally stable~\cite{huttenlocher1979synaptic,han2022adaptive}. 

The pruning functions are updated at each epoch, then we prune neurons with the pruning function $P_i^l<0$. We structurally prune channels in the convolutional layer and prune neurons in the fully connected layer, removing their input connections and output connections. 

\subsection{SNNs Information Transmission}
Different from DNNs, SNNs use spiking neurons with discrete 0/1 output, which are able to integrate spatio-temporal information. Specifically, we employ the leaky integrate-and-fire (LIF) neuron model~\cite{abbott1999lapicque} to transmit and memorize information.
In the spatial dimension, LIF neurons integrate the output of neurons in the previous layer through input synapses. In the temporal dimension, LIF neurons accumulate membrane potentials from previous time steps via internal decay constants $\tau$. Incorporating the spatio-temporal information, the LIF neuron membrane potential $U_i^{t,l}$ at time step $t$ is updated by the following equation:
\begin{equation}
	\label{u}
	U_i^{t,l} =\tau (1-U_i^{t-1,l})+\sum_{j=1}^{M_{l-1}} W_{ij} O_j^{t,l-1}
\end{equation}
When the neuronal membrane potential exceeds the firing threshold $V_{th}$, the neuron fires spike, and its output $O_i^{t,l}$ is equal to 1; Conversely, the neuron outputs 0. The discrete spiking outputs of LIF neurons conserve consumption as the biological brain, but hinder gradient-based backpropagation. To address this problem, \cite{wu2018spatio} first proposed the method of surrogate gradient. In this paper, we use Qgategrad~\cite{qin2020forward} surrogate gradient method with constant $\lambda=2$ to approximate the spiking gradient, as follows:
\begin{equation}
	\label{o}
    \frac{O_i^{t,l}}{U_i^{t,l}}=
        \begin{cases}
        0, & |U_i^{t,l}| > \frac{1}{\lambda} \\
        -\lambda ^2|U_i^{t,l}|+\lambda, & |U_i^{t,l}| \leq \frac{1}{\lambda}
        \end{cases}
\end{equation}
Overall, We present the specific procedure of our DSD-SNN algorithm as Algorithm \ref{alg2}. 
\begin{algorithm}[t]
	\caption{The DSD-SNN Continual Learning.}
	\label{alg2}
	\KwIn{Dataset $D_{T_i}$ for each task $T_i$; \\
	Initialize empty network $Net$;\\
	Constant parameters of growth $\rho\%$ and pruning $\rho_c,\rho_f$.}
	\KwOut{Prediction Class in task $T_i$ (TIL) or in all tasks (CIL).}
    \For{each sequential task $T_i$}{
        Growing new neurons to $Net$ as Eq. \ref{g};\\
	\For{$epoch=0$; $epoch<E$; $epoch++$}{
		SNN forward prediction $Net\left ( D_{T_i} \right ) $ as Eq. \ref{u};\\
		SNN backpropagation to update new connections as Eq. \ref{o};\\
        Assessing importance for newly grown neurons as Eq. \ref{s};\\
	Calculating the neuronal pruning function as Eq. \ref{ddd} and Eq. \ref{fd};	\\
        Pruning redundant neurons with $P_i^l<0$;\\
		}
        Freezing retained neurons in $Net$;\\
        }
	
\end{algorithm}	

\section{Experiments}
\subsection{Datasets and Models}
To validate the effectiveness of our DSD-SNN algorithm, we conduct extensive experiments and analyses on the spatial MNIST~\cite{lecun1998mnist}, CIFAR100~\cite{krizhevsky2009learning} and neuromorphic temporal N-MNIST datasets~\cite{orchard2015converting} based on the brain-inspired cognitive intelligence engine BrainCog~\cite{zeng2022braincog}. The specific experimental datasets and models are as follows:
\begin{itemize}

    \item [$\bullet$]  Permuted MNIST: We permute the MNIST handwritten digit dataset to ten tasks via random permutations of the pixels. Each task contains ten classes, divided into 60,000 training samples and 10,000 test samples. As for the SNN model, we use the SNN with two convolutional layers, one fully-connected layer, and the multi-headed output layer.
 
    \item [$\bullet$] Permuted N-MNIST: We randomly permute the N-MNIST ( the
    neuromorphic capture of MNIST) to ten tasks.  And we employ the same sample division and the same SNN structure as MNIST.
 
    \item [$\bullet$] Split CIFAR100: The more complex natural image dataset CIFAR100 is trained in several splits including 10 steps (10 new classes per step), 20 steps (5 new classes per step).  SNN model consisting of eight convolutional layers, one fully connected and multi-headed output layer are used to generate the predicted class.
    
\end{itemize}

For the task classifier, we use networks containing a hidden layer with 100 hidden neurons for MNIST and N-MNIST, and 500 hidden neurons for CIFAR100. To recognize tasks better, we replay 2000 samples for each task as~\cite{rebuffi2017icarl,rajasegaran2019adaptive,rajasegaran2019random}. Our code is available at \href{https://github.com/BrainCog-X/Brain-Cog/tree/main/examples/Structural_Development/DSD-SNN}{https://github.com/BrainCog-X/Brain-Cog/tree/main/examples/Structural\_Development/DSD-SNN}.

\subsection{Comparisons of Performance}
\begin{figure}[htp] 
	\centering  
	\includegraphics[width=1\linewidth]{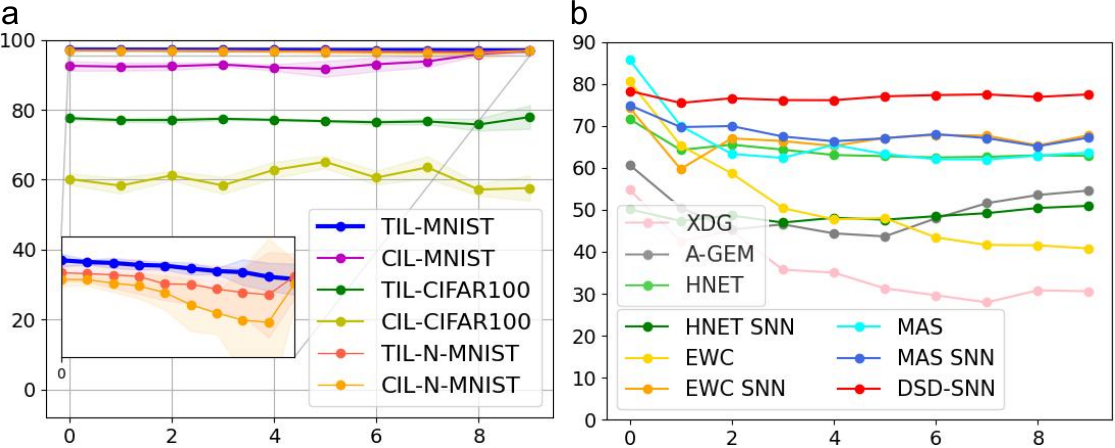} 
	\caption{The average accuracy with increasing number of tasks. \textbf{(a)} Our DSD-SNN for MNIST, N-MNIST and CIFAR100. \textbf{(b)} Comparison of our DSD-SNN with other methods for CIFAR100.}
	\label{tu4}
\end{figure}

As shown in Fig. \ref{tu4}a, our DSD-SNN model maintains high accuracy with increasing number of learned tasks. This demonstrates that the proposed model overcomes catastrophic forgetting on all MNIST, neuromorphic N-MNIST and more complex CIFAR100 datasets, achieving robustness and generalization capability on both TIL and CIL. To validate the effectiveness of our dynamic structure development module, we compare the learning process of DSD-SNN with other DNNs-based continual learning and transfer them to SNN as Fig. \ref{tu4}b. The experimental results indicate that DSD-SNN realizes superior performance in learning and memorizing more incremental tasks, exhibiting larger memory capacity compared to the DNNs-based continual learning baselines.

The comparison results of the average accuracy with existing continual learning algorithms based on DNN and SNN are shown in Table \ref{mnist} and Table \ref{cifar}. In the TIL scenario, our DSD-SNN achieves an accuracy of 97.30\% $\pm$ 0.09\% with a network parameter compression rate of 34.38\% for MNIST, which outperforms most DNNs-based algorithms such as EWC~\cite{kirkpatrick2017overcoming}, GEM~\cite{lopez2017gradient}, and RCL~\cite{xu2018reinforced}. In particular, our algorithm achieves a higher performance improvement of 0.70\% over the DEN~\cite{yoon2017lifelong} model (which is also based on growth and pruning). For the temporal neuromorphic N-MNIST dataset, our DSD-SNN algorithm is superior to the existing HMN algorithm which combines SNN with DNN~\cite{zhao2022framework}. Meanwhile, our DSD-SNN model achieves 92.69\% $\pm$ 0.57\% and 96.94\% $\pm$ 0.05\% accuracy in CIL scenarios for MNIST and N-MNIST, respectively.

\begin{table}[htbp]
  \centering
  \resizebox{3.3in}{!}{
    \begin{tabular}{ccc}
    \toprule
    Method &  Dataset & Acc \\
    \midrule
     EWC~\cite{kirkpatrick2017overcoming}  &  MNIST & 81.60\% \\
     GEM~\cite{lopez2017gradient}  &  MNIST & 92.00\% \\
     DEN~\cite{yoon2017lifelong}  &  MNIST & 96.60\% \\
     RCL~\cite{xu2018reinforced} &  MNIST & 96.60\% \\
     CLNP~\cite{golkar2019continual}  &  MNIST & 98.42\% $\pm$ 0.04 \% \\
     \textbf{Our DSD-SNN} &  MNIST & \textbf{97.30\% $\pm$ 0.09 \%} \\ 
    HMN(SNN+DNN)~\cite{zhao2022framework}  & N-MNIST & 78.18\%\\
    \textbf{Our DSD-SNN}  & N-MNIST & \textbf{97.06\% $\pm$ 0.09 \%} \\
    \bottomrule
    \end{tabular}%
    }
  \caption{Accuracy of task incremental learning compared to other works for MNIST and N-MNIST datasets.}
  \label{mnist}%
\end{table}%

 From Table \ref{cifar}, our DSD-SNN outperforms PathNet~\cite{fernando2017pathnet}, DEN~\cite{yoon2017lifelong}, RCL~\cite{xu2018reinforced} and HNET~\cite{von2020continual}, which are also structural extension methods, in both TIL and CIL scenarios for 10 steps CIFAR100. iCaRL~\cite{rebuffi2017icarl} and DER++~\cite{yan2021dynamically} achieve higher accuracy of 84.20\% in TIL scenarios than our 77.92\%, but they are inferior in CIL scenarios (51.40\% and 55.30\%) than our 60.47\%. Moreover, the DSD-SNN compresses the network to only 37.48\% after learning all tasks, further saving energy consumption. For 20 steps CIFAR100 with more tasks, our DSD-SNN achieves even higher accuracy 81.17\% in TIL scenario and has excellent experimental results consistent with 10 steps. To the best of our knowledge, this is the first time that the energy-efficient deep SNNs have been used to solve CIFAR100 continual learning and achieve comparable performance with DNNs.

\begin{table*}[htbp]
  \centering
  \resizebox{6in}{!}{
    \begin{tabular}{ccccc}
    \toprule
    Method & 10steps TIL Acc (\%) & 10steps CIL Acc (\%) & 20steps TIL Acc (\%) & 20steps CIL Acc (\%) \\
    \midrule
        EWC~\cite{kirkpatrick2017overcoming}  & 61.11 $\pm$ 1.43 & 17.25 $\pm$ 0.09 & 50.04 $\pm$ 4.26 &  4.63 $\pm$ 0.04\\
    MAS~\cite{aljundi2018memory}  & 64.77 $\pm$ 0.78  & 17.07 $\pm$ 0.12 & 60.40 $\pm$ 1.74 & 4.66 $\pm$ 0.02 \\
    PathNet~\cite{fernando2017pathnet}  & 53.10  & 18.50 &-&- \\
    SI~\cite{zenke2017continual} & 64.81 $\pm$ 1.00  & 17.26 $\pm$ 0.11 & 61.10 $\pm$ 0.82 & 4.63 $\pm$ 0.04\\
    DEN~\cite{yoon2017lifelong}  & 58.10  & -&-&-\\
    RCL~\cite{xu2018reinforced}  & 59.90  & -&-&-\\
    iCaRL~\cite{rebuffi2017icarl}  & 84.20 $\pm$ 1.04  & 51.40 $\pm$ 0.99 & 85.70 $\pm$ 0.68 & 47.80 $\pm$ 0.48 \\
    HNET~\cite{von2020continual} &63.57 $\pm$ 1.03 &-& 70.48 $\pm$ 0.25 &-\\
    DER++~\cite{yan2021dynamically}  & 84.20 $\pm$ 0.47  & 55.30 $\pm$ 0.10 & 86.60 $\pm$ 0.50 & 46.60 $\pm$ 1.44\\
    FOSTER~\cite{wang2022foster}  &  - & 72.90 &  - & 70.65 \\
    DyTox~\cite{douillard2022dytox}  &  - & 73.66 $\pm$ 0.02  &  - & 72.27 $\pm$ 0.18\\
    \textbf{Our DSD-SNN}  &  \textbf{77.92 $\pm$ 0.29} &  \textbf{60.47 $\pm$ 0.72} & \textbf{81.17 $\pm$ 0.73 } & \textbf{57.39 $\pm$ 1.97 }\\
    \bottomrule
    \end{tabular}%
    }
  \caption{Accuracy comparisons with DNNs-based algorithms for CIFAR100.}
  \label{cifar}%
\end{table*}%

In summary, the DSD-SNN model significantly outperforms the SNNs-based continual learning model on the N-MNIST dataset. On MNIST and CIFAR100 datasets, the proposed model achieves comparable performance with DNNs-based models and performs well on both TIL and CIL.

\subsection{Effects of Efficient Continual Learning}
\begin{figure}[!htp] 
	\centering  
	\includegraphics[width=1\linewidth]{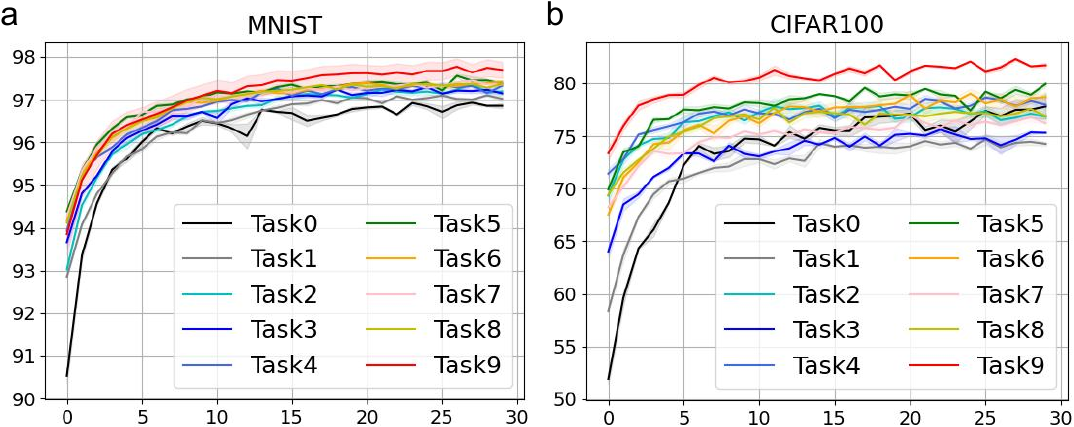} 
	\caption{During the continual learning process of each task, the changes of accuracy with epochs.}
	\label{tu6}
\end{figure}

Fig. \ref{tu6} depicts the performance of the DSD-SNN model for task incremental learning on multiple datasets. The experimental results demonstrate that our SNNs-based model could improve the convergence speed and performance of new tasks during sequential continual learning, possessing the forward transfer capability. The newer tasks achieve higher performance from the beginning for MNIST and CIFAR100 datasets, indicating that the previously learned knowledge is fully utilized to help the new tasks. Also, the new tasks converge to higher performance faster, suggesting that the network has a strong memory capacity to continuously learn and remember new tasks. Similar comparable results can be obtained on the N-MNIST dataset. 

\subsection{Ablation Studies}

\begin{figure}[htp] 
	\centering  
	\includegraphics[width=1\linewidth]{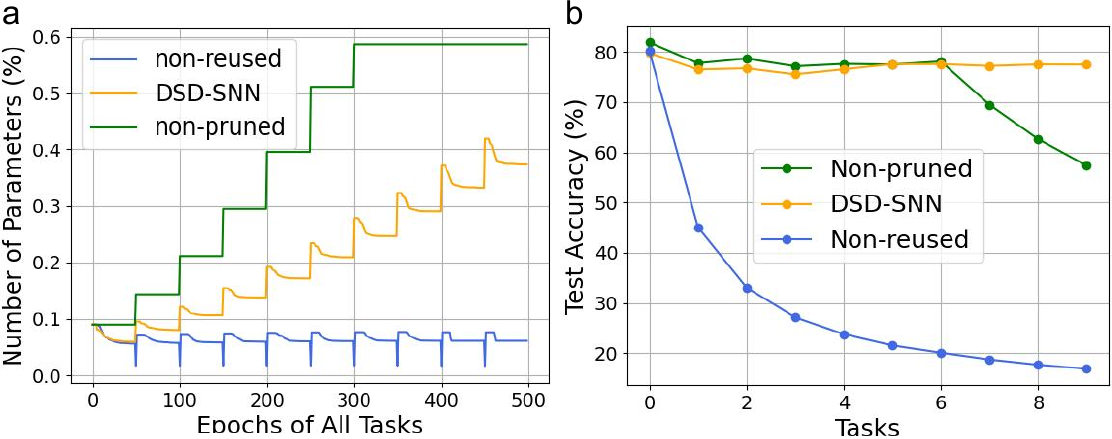} 
	\caption{Effects of each component.  Number of network parameters \textbf{(a)} and accuracy \textbf{(b)} of our DSD-SNN, non-pruned model and non-reused model for CIFAR100.}
	\label{tu5}
\end{figure}

\textbf{Effects of each component.} To verify the effectiveness of the growth and pruning components in our DSD-SNN model, we compare the number of network parameters (Fig.\ref{tu5}a) and performance (Fig.\ref{tu5}b) of DSD-SNN, DSD-SNN without pruning, and DSD-SNN without reused growth during multi-task continual learning. The experimental results show that the number of parameters in the DSD-SNN model fluctuates up and finally stabilizes at 37.48\% for CIFAR100, achieving superior accuracy on multi-task continual learning. In contrast, the network scale of the model without pruning rises rapidly and quickly fills up the memory capacity, leading to a dramatic drop in performance after learning six tasks. The above results reveal that the pruning process of DSD-SNN not only reduces the computational overhead but also improves the performance and memory capacity. 

For the growth module of DSD-SNN, we eliminate the addition of connections from frozen neurons to verify the effectiveness of reusing acquired knowledge in improving learning for new tasks. From Fig. \ref{tu5}a and b, 
DSD-SNN without reused growth suffers from catastrophic forgetting when there is no additional conservation of sub-network masks. The scale of the non-reused network is very small, and the failure to reuse acquired knowledge significantly degrades the performance of the model on each task. Therefore, we can conclude that reusing and sharing acquired knowledge in our DSD-SNN model achieves excellent forward transfer capability.

\begin{figure}[htp] 
	\centering  
	\includegraphics[width=1\linewidth]{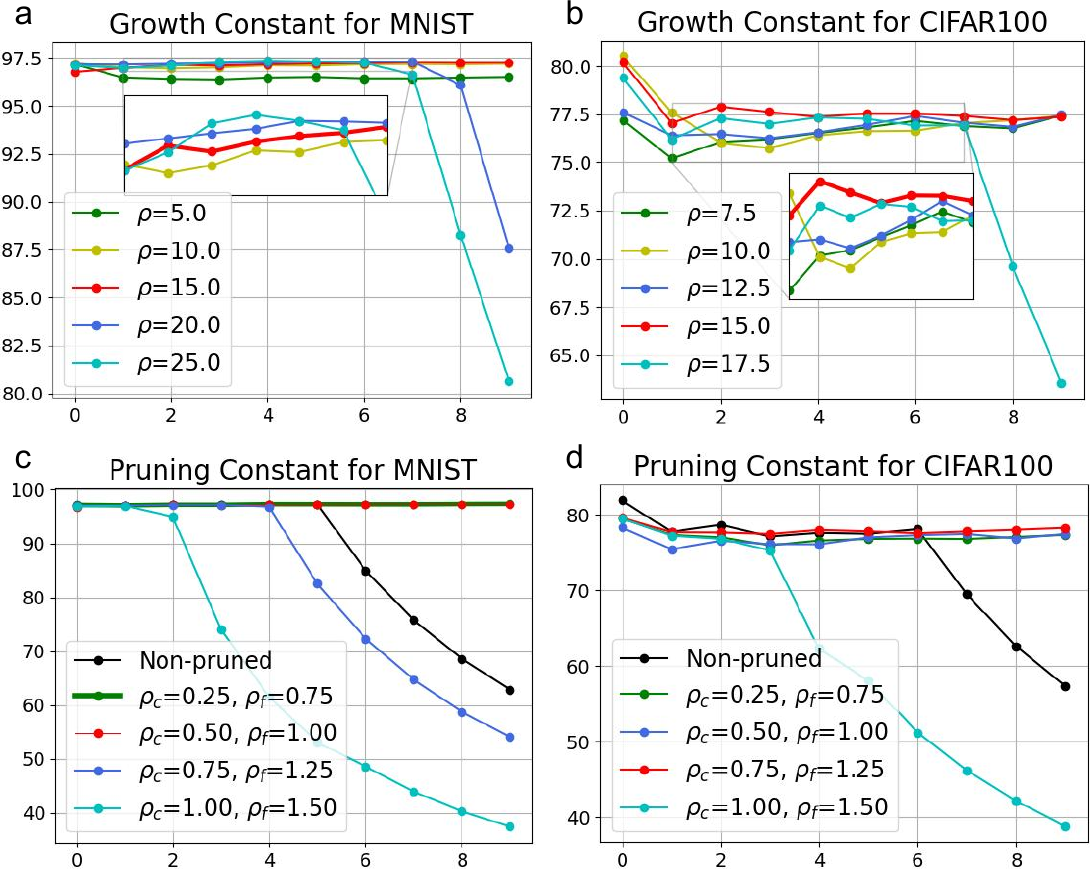} 
	\caption{The effect of pruning and growth parameters on accuracy in multi-task continual learning.}
	\label{tu7}
\end{figure}

\textbf{Effects of different parameters.}
We analyze the effects of different growing and pruning parameters (the growth scale $\rho$ and pruning intensity $\rho_c,\rho_f$). For the growth parameter $\rho$, the results are very close in the range of 5-15\% for MNIST in Fig. \ref{tu7}a, as well as in the range of 7.5-15\% for CIFAR100 in Fig. \ref{tu7}b. Only in the larger case, there is a performance degradation in the later learning task (8th task), due to the larger growth scale of the previous task resulting in insufficient space to learn new knowledge in the later tasks.

Fig. \ref{tu7}c and d describe the effects of pruning strength $\rho_c$, $\rho_f$ on performance. The larger $\rho_c,\rho_f$, the more convolutional channels and fully connected neurons are pruned. We found that the accuracy is very stable at less than $\rho_c=0.50,\rho_f=1.00$ for MNIST and $\rho_c=0.75,\rho_f=1.25$ for CIFAR100, but the accuracy declines at larger $\rho_c,\rho_f$ due to the over-pruning. The DSD-SNN model is more adaptable to pruning parameters on the CIFAR100 dataset because it has a larger parameter space of SNN model. These ablation experiments demonstrate that our DSD-SNN is very robust for different growth and pruning parameters across multiple datasets.

\section{Conclusion}
Inspired by the brain development mechanism, we propose a DSD-SNN model based on dynamic growth and pruning to enhance efficient continual learning. Applied to both TIL and CIL scenarios based on the deep SNN, the proposed model can fully reuse the acquired knowledge to help improve the performance and learning speed of new tasks, and combine with pruning mechanism to significantly reduce the computational overhead and enhance the memory capacity. Our DSD-SNN model belongs to the very few explorations on SNNs-based continual learning. The proposed algorithm surpasses the SOTA performance achieved by SNNs-based continual learning algorithm and achieves comparable performance with DNNs-based continual learning algorithms.

\newpage

\section*{Acknowledgements}
This work is supported by the National Key Research and Development Program (Grant No. 2020AAA0107800), the Strategic Priority Research Program of the Chinese Academy of Sciences (Grant No. XDB32070100),  the National Natural Science Foundation of China (Grant No. 62106261).

\section*{Contribution Statement}
B.H. and F.Z are equal contribution and serve as co-first authors. B.H., F.Z. and Y.Z. designed the study. B.H., F.Z. W.P. and G.S.performed the experiments and the analyses. B.H., F.Z. and Y.Z. wrote the paper.
\bibliographystyle{named}
\bibliography{ijcai23}

\begin{thebibliography}{}

\bibitem[\protect\citeauthoryear{Abbott}{1999}]{abbott1999lapicque}
Larry~F Abbott.
\newblock Lapicque’s introduction of the integrate-and-fire model neuron
  (1907).
\newblock {\em Brain research bulletin}, 50(5-6):303--304, 1999.

\bibitem[\protect\citeauthoryear{Aljundi \bgroup \em et al.\egroup
  }{2018}]{aljundi2018memory}
Rahaf Aljundi, Francesca Babiloni, Mohamed Elhoseiny, Marcus Rohrbach, and
  Tinne Tuytelaars.
\newblock Memory aware synapses: Learning what (not) to forget.
\newblock In {\em Proceedings of the European Conference on Computer Vision
  (ECCV)}, pages 139--154, 2018.

\bibitem[\protect\citeauthoryear{Bruer}{1999}]{bruer1999neural}
John~T Bruer.
\newblock Neural connections: Some you use, some you lose.
\newblock {\em The Phi Delta Kappan}, 81(4):264--277, 1999.

\bibitem[\protect\citeauthoryear{Dekhovich \bgroup \em et al.\egroup
  }{2023}]{dekhovich2022continual}
Aleksandr Dekhovich, David~MJ Tax, Marcel~HF Sluiter, and Miguel~A Bessa.
\newblock Continual prune-and-select: class-incremental learning with
  specialized subnetworks.
\newblock {\em Applied Intelligence}, pages 1--16, 2023.

\bibitem[\protect\citeauthoryear{Douillard \bgroup \em et al.\egroup
  }{2022}]{douillard2022dytox}
Arthur Douillard, Alexandre Ram{\'e}, Guillaume Couairon, and Matthieu Cord.
\newblock Dytox: Transformers for continual learning with dynamic token
  expansion.
\newblock In {\em Proceedings of the IEEE/CVF Conference on Computer Vision and
  Pattern Recognition}, pages 9285--9295, 2022.

\bibitem[\protect\citeauthoryear{Elman \bgroup \em et al.\egroup
  }{1996}]{elman1996rethinking}
Jeffrey~L Elman, Elizabeth~A Bates, and Mark~H Johnson.
\newblock {\em Rethinking innateness: A connectionist perspective on
  development}, volume~10.
\newblock MIT press, 1996.

\bibitem[\protect\citeauthoryear{Fernando \bgroup \em et al.\egroup
  }{2017}]{fernando2017pathnet}
Chrisantha Fernando, Dylan Banarse, Charles Blundell, Yori Zwols, David Ha,
  Andrei~A Rusu, Alexander Pritzel, and Daan Wierstra.
\newblock Pathnet: Evolution channels gradient descent in super neural
  networks.
\newblock {\em arXiv preprint arXiv:1701.08734}, 2017.

\bibitem[\protect\citeauthoryear{French}{1999}]{french1999catastrophic}
Robert~M French.
\newblock Catastrophic forgetting in connectionist networks.
\newblock {\em Trends in cognitive sciences}, 3(4):128--135, 1999.

\bibitem[\protect\citeauthoryear{Furber \bgroup \em et al.\egroup
  }{1987}]{furber1987naturally}
Susan Furber, Ronald~W Oppenheim, and David Prevette.
\newblock Naturally-occurring neuron death in the ciliary ganglion of the chick
  embryo following removal of preganglionic input: evidence for the role of
  afferents in ganglion cell survival.
\newblock {\em Journal of Neuroscience}, 7(6):1816--1832, 1987.

\bibitem[\protect\citeauthoryear{Gao \bgroup \em et al.\egroup
  }{2022}]{gao2022efficient}
Qiang Gao, Zhipeng Luo, Diego Klabjan, and Fengli Zhang.
\newblock Efficient architecture search for continual learning.
\newblock {\em IEEE Transactions on Neural Networks and Learning Systems},
  2022.

\bibitem[\protect\citeauthoryear{Golkar \bgroup \em et al.\egroup
  }{2020}]{golkar2019continual}
Siavash Golkar, Michael Kagan, and Kyunghyun Cho.
\newblock Continual learning via neural pruning.
\newblock In {\em International Conference on Learning Representations}, 2020.

\bibitem[\protect\citeauthoryear{Han \bgroup \em et al.\egroup
  }{2022a}]{han2022adaptive}
Bing Han, Feifei Zhao, Yi~Zeng, and Wenxuan Pan.
\newblock Adaptive sparse structure development with pruning and regeneration
  for spiking neural networks.
\newblock {\em arXiv preprint arXiv:2211.12219}, 2022.

\bibitem[\protect\citeauthoryear{Han \bgroup \em et al.\egroup
  }{2022b}]{han2022developmental}
Bing Han, Feifei Zhao, Yi~Zeng, and Guobin Shen.
\newblock Developmental plasticity-inspired adaptive pruning for deep spiking
  and artificial neural networks.
\newblock {\em arXiv preprint arXiv:2211.12714}, 2022.

\bibitem[\protect\citeauthoryear{Huttenlocher and
  others}{1979}]{huttenlocher1979synaptic}
Peter~R Huttenlocher et~al.
\newblock Synaptic density in human frontal cortex-developmental changes and
  effects of aging.
\newblock {\em Brain Res}, 163(2):195--205, 1979.

\bibitem[\protect\citeauthoryear{Huttenlocher}{1990}]{huttenlocher1990morphometric}
Peter~R Huttenlocher.
\newblock Morphometric study of human cerebral cortex development.
\newblock {\em Neuropsychologia}, 28(6):517--527, 1990.

\bibitem[\protect\citeauthoryear{Jun and Jin}{2007}]{jun2007development}
Joseph~K Jun and Dezhe~Z Jin.
\newblock Development of neural circuitry for precise temporal sequences
  through spontaneous activity, axon remodeling, and synaptic plasticity.
\newblock {\em PloS one}, 2(8):e723, 2007.

\bibitem[\protect\citeauthoryear{Kemker and Kanan}{2018}]{kemker2017fearnet}
Ronald Kemker and Christopher Kanan.
\newblock Fearnet: Brain-inspired model for incremental learning.
\newblock In {\em International Conference on Learning Representations}, 2018.

\bibitem[\protect\citeauthoryear{Kirkpatrick \bgroup \em et al.\egroup
  }{2017}]{kirkpatrick2017overcoming}
James Kirkpatrick, Razvan Pascanu, Neil Rabinowitz, Joel Veness, Guillaume
  Desjardins, Andrei~A Rusu, Kieran Milan, John Quan, Tiago Ramalho, Agnieszka
  Grabska-Barwinska, et~al.
\newblock Overcoming catastrophic forgetting in neural networks.
\newblock {\em Proceedings of the national academy of sciences},
  114(13):3521--3526, 2017.

\bibitem[\protect\citeauthoryear{LeCun \bgroup \em et al.\egroup
  }{1998}]{lecun1998mnist}
Yann LeCun, L{\'e}on Bottou, Yoshua Bengio, and Patrick Haffner.
\newblock Gradient-based learning applied to document recognition.
\newblock {\em Proceedings of the IEEE}, 86(11):2278--2324, 1998.

\bibitem[\protect\citeauthoryear{Li and Hoiem}{2017}]{li2017learning}
Zhizhong Li and Derek Hoiem.
\newblock Learning without forgetting.
\newblock {\em IEEE transactions on pattern analysis and machine intelligence},
  40(12):2935--2947, 2017.

\bibitem[\protect\citeauthoryear{Lopez-Paz and
  Ranzato}{2017}]{lopez2017gradient}
David Lopez-Paz and Marc'Aurelio Ranzato.
\newblock Gradient episodic memory for continual learning.
\newblock {\em Advances in neural information processing systems}, 30, 2017.

\bibitem[\protect\citeauthoryear{Maass}{1997}]{Maass1997Networks}
Wolfgang Maass.
\newblock Networks of spiking neurons: The third generation of neural network
  models.
\newblock {\em Neural networks}, 10(9):1659--1671, 1997.

\bibitem[\protect\citeauthoryear{Orchard \bgroup \em et al.\egroup
  }{2015}]{orchard2015converting}
Garrick Orchard, Ajinkya Jayawant, Gregory~K Cohen, and Nitish Thakor.
\newblock Converting static image datasets to spiking neuromorphic datasets
  using saccades.
\newblock {\em Frontiers in neuroscience}, 9:437, 2015.

\bibitem[\protect\citeauthoryear{Qin \bgroup \em et al.\egroup
  }{2020}]{qin2020forward}
Haotong Qin, Ruihao Gong, Xianglong Liu, Mingzhu Shen, Ziran Wei, Fengwei Yu,
  and Jingkuan Song.
\newblock Forward and backward information retention for accurate binary neural
  networks.
\newblock In {\em Proceedings of the IEEE/CVF conference on computer vision and
  pattern recognition}, pages 2250--2259, 2020.

\bibitem[\protect\citeauthoryear{Rajasegaran \bgroup \em et al.\egroup
  }{2019a}]{rajasegaran2019random}
Jathushan Rajasegaran, Munawar Hayat, Salman Khan, Fahad~Shahbaz Khan, and Ling
  Shao.
\newblock Random path selection for incremental learning.
\newblock {\em Advances in Neural Information Processing Systems}, 3, 2019.

\bibitem[\protect\citeauthoryear{Rajasegaran \bgroup \em et al.\egroup
  }{2019b}]{rajasegaran2019adaptive}
Jathushan Rajasegaran, Munawar Hayat, Salman~H Khan, Fahad~Shahbaz Khan, and
  Ling Shao.
\newblock Random path selection for continual learning.
\newblock {\em Advances in Neural Information Processing Systems}, 32, 2019.

\bibitem[\protect\citeauthoryear{Rebuffi \bgroup \em et al.\egroup
  }{2017}]{rebuffi2017icarl}
Sylvestre-Alvise Rebuffi, Alexander Kolesnikov, Georg Sperl, and Christoph~H
  Lampert.
\newblock icarl: Incremental classifier and representation learning.
\newblock In {\em Proceedings of the IEEE conference on Computer Vision and
  Pattern Recognition}, pages 2001--2010, 2017.

\bibitem[\protect\citeauthoryear{Rusu \bgroup \em et al.\egroup
  }{2016}]{rusu2016progressive}
Andrei~A Rusu, Neil~C Rabinowitz, Guillaume Desjardins, Hubert Soyer, James
  Kirkpatrick, Koray Kavukcuoglu, Razvan Pascanu, and Raia Hadsell.
\newblock Progressive neural networks.
\newblock In {\em In Proceedings of Conference on Neural Information Processing
  Systems}, 2016.

\bibitem[\protect\citeauthoryear{Siddiqui and
  Park}{2021}]{siddiqui2021progressive}
Zahid~Ali Siddiqui and Unsang Park.
\newblock Progressive convolutional neural network for incremental learning.
\newblock {\em Electronics}, 10(16):1879, 2021.

\bibitem[\protect\citeauthoryear{Silva \bgroup \em et al.\egroup
  }{2009}]{silva2009molecular}
Alcino~J Silva, Yu~Zhou, Thomas Rogerson, Justin Shobe, and J~Balaji.
\newblock Molecular and cellular approaches to memory allocation in neural
  circuits.
\newblock {\em Science}, 326(5951):391--395, 2009.

\bibitem[\protect\citeauthoryear{van~de Ven \bgroup \em et al.\egroup
  }{2020}]{van2020brain}
Gido~M van~de Ven, Hava~T Siegelmann, and Andreas~S Tolias.
\newblock Brain-inspired replay for continual learning with artificial neural
  networks.
\newblock {\em Nature communications}, 11(1):1--14, 2020.

\bibitem[\protect\citeauthoryear{Von~Oswald \bgroup \em et al.\egroup
  }{2020}]{von2020continual}
Johannes Von~Oswald, Christian Henning, Benjamin~F Grewe, and Jo{\~a}o
  Sacramento.
\newblock Continual learning with hypernetworks.
\newblock In {\em International Conference on Learning Representations}, 2020.

\bibitem[\protect\citeauthoryear{Wang \bgroup \em et al.\egroup
  }{2022}]{wang2022foster}
Fu-Yun Wang, Da-Wei Zhou, Han-Jia Ye, and De-Chuan Zhan.
\newblock Foster: Feature boosting and compression for class-incremental
  learning.
\newblock In {\em Computer Vision--ECCV 2022: 17th European Conference, Tel
  Aviv, Israel, October 23--27, 2022, Proceedings, Part XXV}, pages 398--414.
  Springer, 2022.

\bibitem[\protect\citeauthoryear{Wu \bgroup \em et al.\egroup
  }{2018}]{wu2018spatio}
Yujie Wu, Lei Deng, Guoqi Li, Jun Zhu, and Luping Shi.
\newblock Spatio-temporal backpropagation for training high-performance spiking
  neural networks.
\newblock {\em Frontiers in neuroscience}, 12:331, 2018.

\bibitem[\protect\citeauthoryear{Xu and Zhu}{2018}]{xu2018reinforced}
Ju~Xu and Zhanxing Zhu.
\newblock Reinforced continual learning.
\newblock {\em Advances in Neural Information Processing Systems}, 31, 2018.

\bibitem[\protect\citeauthoryear{Xu \bgroup \em et al.\egroup
  }{2015}]{krizhevsky2009learning}
Bing Xu, Naiyan Wang, Tianqi Chen, and Mu~Li.
\newblock Empirical evaluation of rectified activations in convolutional
  network.
\newblock {\em arXiv preprint arXiv:1505.00853}, 2015.

\bibitem[\protect\citeauthoryear{Yan \bgroup \em et al.\egroup
  }{2021}]{yan2021dynamically}
Shipeng Yan, Jiangwei Xie, and Xuming He.
\newblock Der: Dynamically expandable representation for class incremental
  learning.
\newblock In {\em Proceedings of the IEEE/CVF Conference on Computer Vision and
  Pattern Recognition}, pages 3014--3023, 2021.

\bibitem[\protect\citeauthoryear{Yoon \bgroup \em et al.\egroup
  }{2018}]{yoon2017lifelong}
Jaehong Yoon, Eunho Yang, Jeongtae Lee, and Sung~Ju Hwang.
\newblock Lifelong learning with dynamically expandable networks.
\newblock In {\em International Conference on Learning Representations}, 2018.

\bibitem[\protect\citeauthoryear{Zeng \bgroup \em et al.\egroup
  }{2022}]{zeng2022braincog}
Yi~Zeng, Dongcheng Zhao, Feifei Zhao, Guobin Shen, Yiting Dong, Enmeng Lu, Qian
  Zhang, Yinqian Sun, Qian Liang, Yuxuan Zhao, et~al.
\newblock Braincog: A spiking neural network based brain-inspired cognitive
  intelligence engine for brain-inspired ai and brain simulation.
\newblock {\em arXiv preprint arXiv:2207.08533}, 2022.

\bibitem[\protect\citeauthoryear{Zenke \bgroup \em et al.\egroup
  }{2017}]{zenke2017continual}
Friedemann Zenke, Ben Poole, and Surya Ganguli.
\newblock Continual learning through synaptic intelligence.
\newblock In {\em International Conference on Machine Learning}, pages
  3987--3995. PMLR, 2017.

\bibitem[\protect\citeauthoryear{Zhao and Zeng}{2021}]{zhao2021dynamically}
Feifei Zhao and Yi~Zeng.
\newblock Dynamically optimizing network structure based on synaptic pruning in
  the brain.
\newblock {\em Frontiers in Systems Neuroscience}, 15:620558, 2021.

\bibitem[\protect\citeauthoryear{Zhao \bgroup \em et al.\egroup
  }{2022a}]{zhao2022toward}
Feifei Zhao, Yi~Zeng, and Jun Bai.
\newblock Toward a brain-inspired developmental neural network based on
  dendritic spine dynamics.
\newblock {\em Neural Computation}, 34(1):172--189, 2022.

\bibitem[\protect\citeauthoryear{Zhao \bgroup \em et al.\egroup
  }{2022b}]{zhao2022nature}
Feifei Zhao, Yi~Zeng, Bing Han, Hongjian Fang, and Zhuoya Zhao.
\newblock Nature-inspired self-organizing collision avoidance for drone swarm
  based on reward-modulated spiking neural network.
\newblock {\em Patterns}, 3(11):100611, 2022.

\bibitem[\protect\citeauthoryear{Zhao \bgroup \em et al.\egroup
  }{2022c}]{zhao2022framework}
Rong Zhao, Zheyu Yang, Hao Zheng, Yujie Wu, Faqiang Liu, Zhenzhi Wu, Lukai Li,
  Feng Chen, Seng Song, Jun Zhu, et~al.
\newblock A framework for the general design and computation of hybrid neural
  networks.
\newblock {\em Nature communications}, 13(1):1--12, 2022.

\end{thebibliography}

\end{document}